\documentclass[runningheads]{llncs}
\usepackage[T1]{fontenc}
\usepackage{graphicx}
\usepackage{amsmath}
\usepackage{amsthm}
\usepackage{amssymb}
\usepackage{amsfonts}
\usepackage{amstext}
\usepackage{multirow}

\begin{document}
\title{PBa-LLM: Privacy- and Bias-aware NLP using Named-Entity Recognition (NER)}

\titlerunning{PBa-LLM: Privacy- and Bias-aware NLP using NER}
\author{
    Gonzalo Mancera\inst{1} \and
    Aythami Morales\inst{1} \and
    Julian Fierrez\inst{1} \and
    Ruben Tolosana\inst{1} \and
    Alejandro Peña\inst{1} \and
    Miguel Lopez-Duran\textsuperscript{\rm 1} \and
    Francisco Jurado\inst{2} \and
    Alvaro Ortigosa\inst{2}
}
\authorrunning{Mancera et al.} 
\institute{
    Biometrics and Data Pattern Analytics Lab, Universidad Autonoma de Madrid, Spain \and
    Department of Computer Engineering, Universidad Autonoma de Madrid, Spain \\
    \email{\{gonzalo.mancera, aythami.morales, julian.fierrez, ruben.tolosana, alejandro.penna, miguel.lopezd\\ francisco.jurado, alvaro.ortigosa\}@uam.es}
}
\maketitle              
\begin{abstract}
 The use of Natural Language Processing (NLP) in high-stakes AI-based applications has increased significantly in recent years, especially since the emergence of Large Language Models (LLMs). However, despite their strong performance, LLMs introduce important legal/ethical concerns, particularly regarding privacy, data protection, and transparency. Due to these concerns, this work explores the use of Named-Entity Recognition (NER) to facilitate the privacy-preserving training (or adaptation) of LLMs. We propose a framework that uses NER technologies to anonymize sensitive information in text data, such as personal identities or geographic locations. An evaluation of the proposed privacy-preserving learning framework was conducted to measure its impact on user privacy and system performance in a particular high-stakes and sensitive setup: AI-based resume scoring for recruitment processes. The study involved two language models (BERT and RoBERTa) and six anonymization algorithms (based on Presidio, FLAIR, BERT, and different versions of GPT) applied to a database of $24$,$000$ candidate profiles. The findings indicate that the proposed privacy preservation techniques effectively maintain system performance while playing a critical role in safeguarding candidate confidentiality, thus promoting trust in the experimented scenario. On top of the proposed privacy-preserving approach, we also experiment applying an existing approach that reduces the gender bias in LLMs, thus finally obtaining our proposed Privacy- and Bias-aware LLMs (PBa-LLMs). Note that the proposed PBa-LLMs have been evaluated in a particular setup (resume scoring), but are generally applicable to any other LLM-based AI application.

\keywords{Privacy \and Bias \and Fairness \and Responsible AI \and NLP \and Large Language Models \and LLM \and Named Entity Recognition \and NER}
\end{abstract}
\section{Introduction}
During the past decade, automated decision-making systems have increasingly permeated sensitive domains such as education \cite{2023_AAAI_edBB-Demo_Daza,becerra2025aibased}, healthcare \cite{ai4food,staab2024large}, finance \cite{serrano2018automatic}, and recruitment \cite{al2021role,2020_CVPRw_FairCVtest_Pena,2020_ICMI_FairDemo_Pena}, driven by rapid advances in AI technologies. As these systems assume tasks traditionally requiring human judgment, they raise critical questions about personal data protection \cite{acquisti2015privacy,diakopoulos2015algorithmic,voigt2017eu,ghafourian2025blockchainbiometricssurveygdpr}. In particular, the field of NLP has seen unprecedented growth, powered by the Transformer model \cite{vaswani2017attention}, which relies on self-attention mechanisms and has enabled the development of powerful LLMs such as GPT \cite{achiam2023gpt}, LLaMA \cite{touvron2023llama,touvron2023llama2} and Mistral \cite{jiang2023mistral}. These models are now embedded in numerous applications, including those that process sensitive information, such as automated resume screening systems \cite{geetha2018recruitment}.

\begin{figure}[t!] \centering \includegraphics[width=0.8\textwidth]{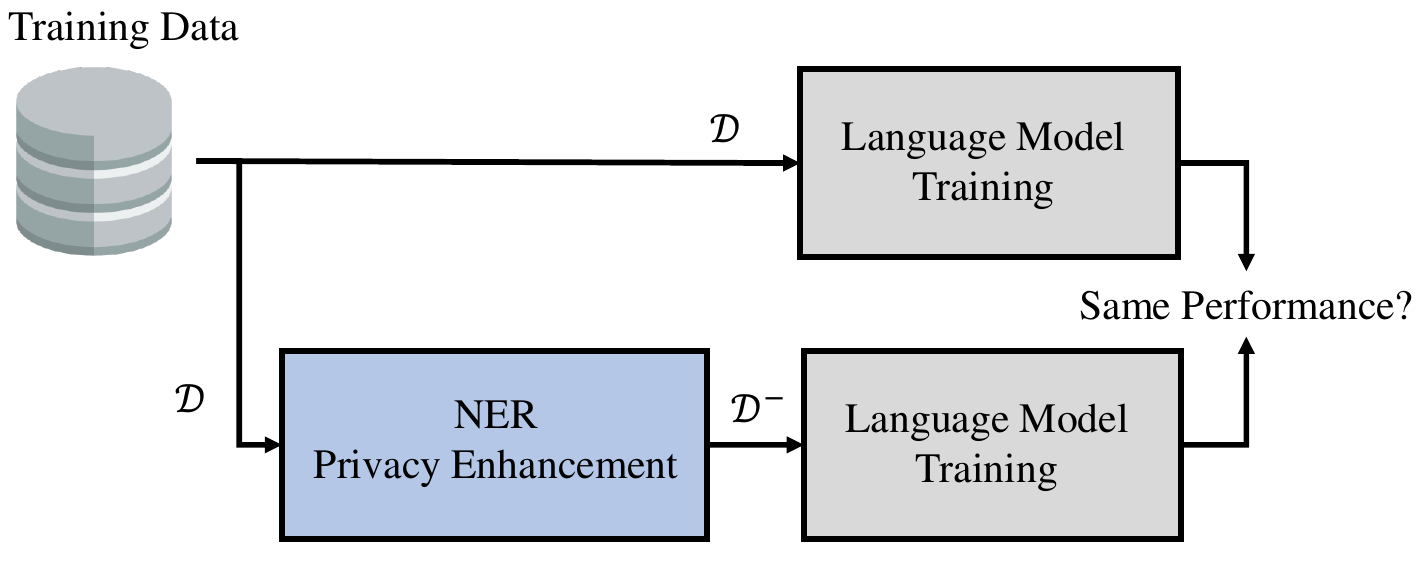} \caption{Graphical abstract of the framework evaluated in this work.} \label{fig:abstract} \end{figure}

Although offering operational advantages, these technologies also face challenges related to fairness \cite{2022_AI_SensitiveLoss_IS,2023_COMPSAC_BiasAI_N-sigma_DeAlcala,peña2025bias} and privacy \cite{cheng2021socially,lepri2021ethical,privacy2025}. Text data often contains personally identifiable information (PII) \cite{schwartz2011pii} - names, addresses, and demographic details that, if mishandled, could expose individuals or introduce bias into decision-making processes. Recent regulatory initiatives, such as the AI Act\footnote{\url{https://artificialintelligenceact.eu/}} approved by the European Parliament in June 2024 \cite{madiega2021artificial} and the White House Memorandum on AI leadership \cite{USA}, seek to address these risks by promoting transparency, accountability, and fairness in AI systems.

The intersection of privacy and fairness in multiple AI applications based on human data (e.g. biometrics and NLP \cite{pena2023human})  has fueled interest in privacy enhancement technologies (PET) for multimodal human-centric AI including NLP modules \cite{kaaniche2020privacy}. Techniques such as differential privacy \cite{2022_Access_DP-CL_Ahmad}, de-identification \cite{2023_COMPSAC_DeID-Adversarial_Mahdi}, crypto-biometrics \cite{2017_Access_HEmultiDTW_Marta}, data minimization \cite{pfitzmann2010terminology}, and federated learning \cite{li2020review,mahdiICIP} are being explored to protect sensitive information while maintaining system performance. Nevertheless, achieving a balance between privacy preservation and model accuracy remains a significant challenge, particularly for complex tasks like natural language understanding.

This paper addresses this challenge by proposing a generalizable approach to anonymize input text while preserving system performance, in a similar way related works remove unnecessary sensitive data while maintaining good performance in a main AI task \cite{morales2020sensitivenets}, in this case particularly applied to NLP tasks based on LLMs (e.g., document layout analysis \cite{PENA-Layout} or document topic classification \cite{2023_ICDAR_LLMs-TopicsPublicDocs_Pena}). Our method uses NER to detect and remove sensitive entities, minimizing privacy risks. The main idea is depicted in Fig.~\ref{fig:abstract}, where we show two branches: with and without removing the sensitive entities found using Named-Entity Recognition. In addition to that privacy enhancement via NER, we also apply and experiment the bias reduction methods recently proposed in \cite{peña2025bias} to derive our final Privacy- and Bias-aware models (PBa-LLMs).

Although recruitment is our primary case study, the framework is designed to extend to other domains, offering a versatile solution for privacy-preserving NLP applications.

The remainder of the paper is organized as follows. Section~\ref{sec:related_works} reviews the related literature on privacy-preserving technologies in NLP. Section~\ref{sec:privacy} details our proposed framework and its integration of NER-based anonymization. Section~\ref{sec:material} describes the experimental setup. Section~\ref{sec:experiments} presents the results and discusses their implications. Finally, Section~\ref{sec:conclusions} summarizes our contributions.

\section{Related Works}
\label{sec:related_works}
Our work falls under the development of privacy-enhancing technologies to handle sensitive data, particularly for NLP applications. Techniques such as \textit{NER} have been widely used to anonymize personal information while maintaining the utility of NLP data for various tasks. Recent advances in \textit{NLP}, including the integration of pre-trained models such as \textit{BERT-} and \textit{GPT}-based models, have further improved entity recognition and anonymization capabilities \cite{staab2024large}.

\subsection{Privacy in NLP Systems}
Data privacy has become an increasingly critical issue in our data-driven society. With exponential growth in the collection and processing of sensitive personal information, the risks of breaches and misuse are more pronounced than ever. Personal data such as names, ages, locations, and organizational affiliations raise significant concerns about acquisition, storage, and use. Mishandling personal data can inadvertently reveal more than intended, leading to discrimination \cite{2022_AI_SensitiveLoss_IS}, identity theft, and other harm. Thus, managing sensitive information, particularly personally identifiable details, requires rigorous measures to ensure confidentiality and integrity.

Traditional privacy protection methods often fail to address the complexities of contemporary data use. Techniques such as pseudonymization, access control, and basic encryption are common. Pseudonymization replaces identifiable data with pseudonyms, but privacy can still be compromised if mappings or contextual clues are accessible \cite{stalla2016anonymous}. Access control restricts sensitive data to authorized individuals but may fail if poorly implemented \cite{sandhu1994access}. Basic encryption is crucial, but its effectiveness is based on strong algorithms and proper key management \cite{neumann2019pseudonymization}. Moreover, machine learning models are vulnerable to adversarial attacks that can expose training data \cite{dealcala2024my,mancera2025my}.

As machine learning and artificial intelligence (AI) technologies continue to advance, the reliance on large datasets presents a double-edged sword. While driving progress, it increases the risk of exposure to sensitive information. Therefore, developing robust privacy-preserving techniques to safeguard sensitive data has become more critical than ever.

\subsection{Named-Entity Recognition}
\textit{Named-Entity Recognition (NER)} is a fundamental task within the broader field of \textit{NLP}, aimed at detecting and categorizing entities in textual data. The primary goal of NER is to identify specific entities within a body of text and classify them into predefined categories \cite{nersurvey}.

The NER process plays a pivotal role in converting unstructured data into structured and meaningful information by automatically extracting relevant entities from large datasets. This is particularly valuable in applications such as Machine Translation \cite{babych2003improving}, Information Retrieval \cite{weston2019named}, Question Answering \cite{molla2006named}, or Sentiment Analysis \cite{JURADOSentiment,ding2018entity}. 

\begin{figure}
\centering
\includegraphics[scale=0.65, angle=-90]{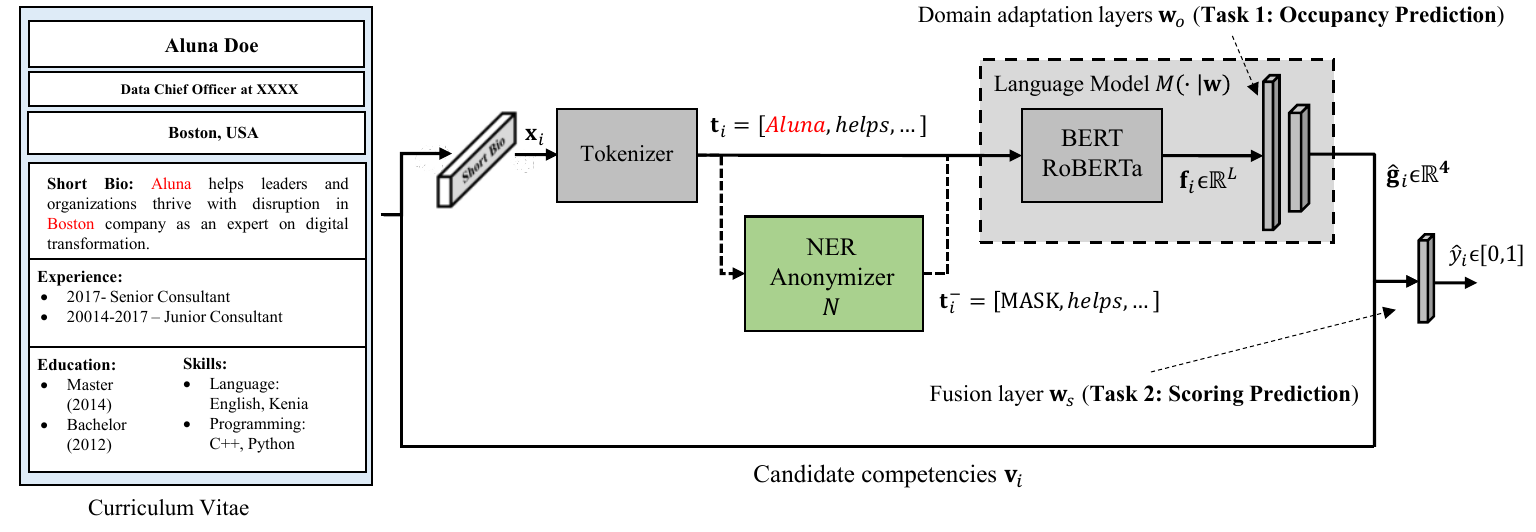}
\caption{Learning framework for the automatic recruitment tool. The system takes a multimodal input resume, consisting of a candidate competencies vector $\mathbf{v}$ and a text biography. An anonymization module $N$ removes two types of identities: locations (LOC) and persons (PER). A pre-trained Transformer processes the anonymized biographies to extract a feature vector $\mathbf{f}$. This feature vector is fused with the candidate competencies using the fusion module $\mathbf{w}_s$ to generate the final score $\hat{y}$.}
\label{fig:architecture}
\end{figure}

Identifying and categorizing entities significantly improves the performance of these applications, making it easier to process and analyze textual data effectively. Beyond these applications, NER serves as a cornerstone in the broader analysis of textual data, enhancing the ability to extract, structure \cite{mumtaz2023extracting}, and interpret information meaningfully \cite{fernandez2011semantically}. By providing a clear framework for understanding entities within text, NER facilitates further analysis, such as trend detection \cite{ritter2011named}, relationship mapping \cite{nasar2021named}, document layout analysis \cite{PENA-Layout}, topic classification \cite{2023_ICDAR_LLMs-TopicsPublicDocs_Pena}, and even predictive analytics \cite{fu2021spanner}.

As the volume of unstructured data continues to grow exponentially in the current digital age, the importance of NER in processing and obtaining insights from these data becomes increasingly critical. Consequently, continued development and refinement of NER technologies remain essential to harness the full potential of big data and drive informed decision making in various sectors.

\section{Privacy-Preserving Learning Approach}
\label{sec:privacy}
In this section, we outline our contribution, which focuses on improving privacy in various tasks. By implementing privacy-preserving techniques, we aim to protect sensitive information across multiple domains. Our approach ensures that personal data remains protected while maintaining the functionality of the tasks, reducing potential exposure to privacy risks without sacrificing performance or utility. This work contributes to ongoing efforts to balance data privacy and task effectiveness, which is critical in an era of increasing concerns about data privacy and individual rights.

\subsection{Problem Formulation} \label{sec:problemformulation}

Consider a language model $M$, defined by its parameters $\mathbf{w}$, trained with a dataset $\mathcal{D}$ to optimize a performance metric $P$. Each data sample $i$ in $\mathcal{D}$ consists of a raw text $\mathbf{x}_i$ and optionally its respective label $\textbf{y}_i$. Text data can be represented as a sequence of $\tau$ tokens $\textbf{t}_i = [t^{(1)}_i, \ldots, t^{(\tau)}_i]$. Assume that there is sensitive information $z_i$ within $\mathbf{t}_i$ that we want to protect. We use the anonymization function $N$ to generate $\mathbf{t}^{-}_i$, so $z_i \notin \textbf{t}^{-}_i$. Based on $N$, we transform $\mathcal{D}$ into a new anonymized version $\mathcal{D}^{-}$ (see Fig. \ref{fig:abstract}). 

The purpose of the proposed privacy-preserving learning framework is to train a model $M^{-}$ with the anonymized dataset $\mathcal{D}^{-}$ to ensure the removal of $z$ while preserving performance, that is, $P(M)=P(M^{-})$.

Specifically, in our experiments we consider a dataset $\mathcal{D}$ of multimodal resumes, where the $i$-th resume consists of a vector of attributes or candidate competencies $\mathbf{v}_i$, along with a biography $\mathbf{x}_i$ tokenized as $\mathbf{t}_i$. In this dataset, sensitive information $z_i$ is only present in biographies (e.g., names, locations, companies). Text data is subsequently processed through the anonymization module $N$, responsible for protecting sensitive data based on \textit{NER} models. Concretely, this module identifies the entities present in $\mathbf{t}$, and produces an anonymized sequence $\mathbf{t}^{-}$ by removing sensitive data, which will typically be personal information of the candidate, embedded inadvertently in the short biography (e.g., name, surname, locations, companies). In this framework, a machine learning-based resume scoring tool is developed based on the general learning architecture illustrated in Figure \ref{fig:architecture}. This system is trained on the dataset $\mathcal{D}$ to accurately estimate a new score $\hat{y}_i$ that approximates a target score $y_i$ (e.g., supervised labels produced by human resources experts) utilizing the input resume $\{\mathbf{v}_i, \mathbf{t}_i\}$. It is important to note that the scoring mechanism comprises: i) a pre-trained Transformer (e.g., BERT and RoBERTa in our experiments), which accepts a tokenized biography as input and generates a text embedding $\mathbf{f}_i \in \mathbb{R}^L $; and ii) a fusion module, which further processes $\textbf{f}_i$ and integrates it with the candidate attributes $\mathbf{v}_i$ to predict the score $\hat{y}_i$.


Based on the experimental frameworks proposed in \cite{pena2023human} we incorporate the analysis of two different learning tasks: \textit{i}) Occupancy prediction; \textit{ii}) Score prediction.


The models were first trained using the original $\mathbf{t}$ to establish a non-private baseline and are subsequently independently trained on $\textbf{t}^{-}$ to assess differences in performance and preservation of privacy. 

\subsection{NER Models}
The NER-based privacy module is designed to automatically anonymize sensitive information in texts using advanced NER techniques. Its primary goal is to identify and replace personal or private data, such as the names, locations, or other entities of individuals that could compromise privacy. This module can be categorized into two main groups: 1) specifically designed for NER tasks, which utilize tailored algorithms and models to accurately identify entities within the text, ensuring high precision and recall for effective detection of sensitive information, and 2) LLMs, which are primarily developed to handle a diverse set of tasks in NLP. Here, we experimentally compare both categories: the use of specialized NER models versus general LLMs for anonymization. The following approaches are utilized:

\begin{itemize}
    \item \textbf{Presidio} \footnote{\url{https://microsoft.github.io/presidio/}} is an open-source tool by \textit{Microsoft} for detecting and anonymizing sensitive information. Its architecture includes an \textit{Analyzer} to detect PII such as names or addresses, and an \textit{Anonymizer} to replace or mask them.

    \item \textbf{FLAIR} \cite{akbik2019flair} is a pre-trained NER model based on a bidirectional LSTM and \textit{contextual string embeddings}, which capture word meaning from the surrounding context, improving disambiguation and sense adaptation.

    \item \textbf{NER\_CoNLL2003\_BERT} is a BERT-based model fine-tuned on the CoNLL-2003 dataset \cite{devlin2018bert}, offering strong generalization and robust entity extraction through bidirectional context representation.

    \item \textbf{GPT-3.5} \cite{ouyang2022instructgpt} is a general purpose Transformer model enhanced with sparse attention and RLHF, enabling better alignment with human intent and better understanding of text.

    \item \textbf{GPT-4} \cite{achiam2023gpt} introduces refined attention mechanisms and fine-tuning for improved contextual performance. The lightweight \textbf{GPT-4o Mini} is also evaluated for low-latency, cost-efficient applications.
\end{itemize}

\section{Experimental Framework}
\label{sec:material}
This section first introduces the object databases used in this study, describing their structure, key characteristics, and relevance to the analysis. Following this, we detail the experimental protocol itself, outlining the methodologies applied, and the experimental setup.

\subsection{Dataset: FairCVdb}
\label{sec:dataset}

FairCVdb \cite{2020_CVPRw_FairCVtest_Pena,2020_ICMI_FairDemo_Pena,pena2023human} is a comprehensive dataset that aims to advance research on multimodal fairness within the realm of artificial intelligence. This dataset comprises $24$,$000$ synthetic profiles generated by integrating identities sourced from the DiveFace Face Recognition database \cite{morales2020sensitivenets} and leveraging biographies. Each profile in FairCVdb includes:
\begin{itemize}
    \item \textbf{Textual Information}: FairCVdb was created using data from the occupation database \cite{de2019bias}, with samples assigned to profiles based on gender. Each sample contains two brief biographies (one raw and one gender neutral), an occupation and a name covering ten occupations in four labor sectors. This diversity in occupations allows for a broader analysis of privacy in several fields.
    \item \textbf{Structured Information}: Seven attributes were derived from five categories that define candidate competencies: education, availability, previous experience, recommendations, and language proficiency. Gender and ethnicity were also included as demographic factors. Each resume received a blind score (excluding demographics) and biased scores, with sector-specific weightings. The candidates were labeled with an occupational group ($g \in$ O1: Health, O2: Education and Knowledge, O3: Communication and Media, O4: Legal) based on their profession and assigned a human-evaluated score $y \in [0, 1]$ following human resource processes \cite{pena2023human}.
\end{itemize}

 FairCVdb includes gender-biased scores generated to assess performance and fairness in automatic recruitment tools. By analyzing the performance of these scores, it is possible to investigate the disparities in hiring and evaluation processes, explore the implications of demographic attributes on employment outcomes, and propose strategies for enhancing fairness and privacy in multimodal AI systems. 

\subsection{Privacy Enhancing Module}
The model $N$ is responsible for the anonymization of the texts according to the desired entity. In the present research work, two distinct entities are differentiated: Location and Person. A total of six distinct models are utilized to observe how the choice of anonymizer can affect the final performance of the tasks. 

All anonymizers receive raw information $\mathbf{t}$, which consists of textual data from each of the 24,000 resumes present in FairCVdb, and the model output retains the same contextual information from each resume, except that the selected entity is masked, resulting in $\mathbf{t}^{-}$.

 The analysis was carried out on the short biographies available in the FairCVdb database, which contains a total of 1,416,693 words. The number of entities removed varies from $28$,$000$ to $66$,$000$ depending on the NER model and the type of entity. This information represents approximately between $4\%$ and $9\%$ of the total words included in the short biographies. The variation in the number of entities identified by the NER algorithms is mainly due to their different performances and different sensitivities. For example, compound nouns may be recognized as single or multiple entities based on the algorithm used. Although the NER algorithms assessed are not without errors, each of the six NER modules achieved performance exceeding $80\%$ on the public \textit{CoNLL-}$2003$ benchmark \cite{sang2003introduction}.

\subsection{Occupancy Classification and Candidate Scoring}
\label{sec:tasks}
We used two distinct pre-trained Transformer models for the Text Comprehension Module in our experiments: \textit{BERT} and \textit{RoBERTa}. 

\begin{itemize}
    \item \textbf{BERT} \cite{devlin2018bert} is a Transformer-based model that introduced bidirectional pre-training using \textit{Masked Language Modeling (MLM)} and \textit{Next Sentence Prediction (NSP)}. It captures left and right context simultaneously, enabling deep understanding of sentence structure and inter-sentence relationships.

    \item \textbf{RoBERTa} \cite{liu2019roberta} improves on BERT by removing the NSP objective, training on a larger corpus and dynamically applying masking at each iteration. These adjustments enhance performance while keeping the original architecture.
\end{itemize}

We employ here the base versions of models, characterized by presenting a latent space embedding of dimensions $768$. To maintain the integrity of their pre-trained weights, we freeze the Transformer layers throughout the training process. The feature vector $\mathbf{f}_i \in \mathbb{R}^{768}$ is generated by averaging the embeddings of the output tokens of the Transformer model, ensuring that the attention mask is applied to exclude padding tokens from the averaging process. Before feeding the bios into the models, we preprocess them by eliminating punctuation and using the respective tokenizer, adhering to a fixed sequence length of $256$, since the longest bio in the training dataset comprises $197$ words.

\begin{itemize}

\item For the \textbf{Occupancy Prediction task} we have trained a \textit{Multilayer Feedforward Perceptron (MLP}) with three layers characterized by their parameters $\mathbf{w}_o$. The number of units in each layer is $300$, $70$, and $4$, respectively. The output of $\mathbf{w}_o$ is a vector $\mathbf{g}_i$ corresponding to the probabilities of the four predefined occupancy sectors in FairCVdb.

\item Subsequently, for the \textbf{Scoring Prediction task}, we concatenate the outputs of these layers with the candidate competencies $\mathbf{v}_i$ to predict the final score $\hat{y}$ through an output layer ($1$ unit) characterized by the parameters $\mathbf{w}_s$.

\end{itemize}

All trained layers ($\mathbf{w}_o$ and $\mathbf{w}_s$) use the sigmoid activation function, with a dropout rate set at $0.3$ to alleviate the risk of overfitting. The layers were trained during $50$ epochs, using the root mean square error (RMSE) as a loss function, with a batch size of $32$. We employed the \textit{AdamW} optimizer with parameters \( \beta_1 = 0.9 \), \( \beta_2 = 0.999 \), and \( \epsilon = 2 \times 10^{-8} \), along with a learning rate of \( 1 \times 10^{-3} \). FairCVdb contains a total of $24$,$000$ synthetic resumes and is divided into training and testing subsets for model evaluation. Specifically, $80\%$ of the data, amounting to $19$,$200$ resumes, is used for training purposes. The remaining $20\%$, comprising $4$,$800$ resumes, is designated for testing, providing an independent set of data to assess the model's generalization capabilities.

\section{Experiments and Results}
\label{sec:experiments}

\begin{table*}[tp]
\centering
\resizebox{\columnwidth}{!}{
\begin{tabular}{llcccccccc}
\hline
\multicolumn{1}{l}{}                                            &                                        & \multicolumn{4}{c}{\textbf{LOC}}                                                                                                                                          & \multicolumn{4}{c}{\textbf{PER}}                                                                                                                                          \\ 
\multicolumn{1}{l}{\textbf{Transformer}} & \textbf{Anonymization Module} & \multicolumn{1}{c}{\textbf{O1}} & \multicolumn{1}{c}{\textbf{O2}} & \multicolumn{1}{c}{\textbf{O3}} & \multicolumn{1}{c}{\textbf{O4}} & \multicolumn{1}{c}{\textbf{O1}} & \multicolumn{1}{c}{\textbf{O2}} & \multicolumn{1}{c}{\textbf{O3}} & \multicolumn{1}{c}{\textbf{O4}} \\ \hline
\multirow{7}{*}{RoBERTa}                                    & \textbf{None (Baseline)}                          & $0.86$                                     & $0.88$                                     & $0.85$                                     & $0.79$                                     & $0.89$                                     & $0.84$                                      & $0.87$                                     & $0.72$\\ 
								& GPT-$4$                                  & $0.92$                                     & $0.94$                                     & $0.87$                                     & $0.80$                                     & $0.91$                                     & $0.87$                                     & $0.87$                                     & $0.79$                                         \\
                                                                & GPT-$4$o-mini                            & $0.91$                                     & $0.93$                                    & $0.86$                                     & $0.78$                                     & $0.91$                                     & $0.84$                                     & $0.85$                                     & $0.66$                                     \\
                                                                & GPT-$3.5$                                & $0.83$                                     & $0.82$                                     & $0.86$                                     & $0.66$                                     & $0.82$                                     & $0.86$                                     & $0.89$                                     & $0.71$                                     \\
                                                                & Presidio                               & $0.91$                                     & $0.89$                                     & $0.88$                                     & $0.79$                                     & $0.93$                                     & $0.85$                                     & $0.88$                                     & $0.69$                                     \\
                                                                & FLAIR                                  & $0.92$                                     & $0.87$                                     & $0.84$                                     & $0.71$                                     & $0.91$                                     & $0.85$                                     & $0.88$                                     & $0.78$                                     \\
                                                                & NER CoNLL$2003$ BERT                     & $0.93$                                     & $0.84$                                     & $0.81$                                     & $0.72$                                     & $0.85$                                     & $0.86$                                     & $0.88$                                     & $0.76$                                     \\ \hline
\multirow{7}{*}{BERT}                                       & \textbf{None (Baseline)}                          & $0.87$                                     & $0.85$                                     & $0.87$                                     & $0.79$                                     & $0.89$                                     & $0.86$                                      & $0.87$                                     & 0.79\\ 
								& GPT-$4$                                  & $0.92$                                         & $0.87$                                          &$0.89$                                          &$0.81$                                          & $0.92$                                          & $0.88$                                         & $0.87$                                         & $0.82$                                         \\
                                                                & GPT-$4$o-mini                            & $0.91$                                     & $0.86$                                     & $0.90$                                     & $0.77$                                     & $0.90$                                     & $0.88$                                     & $0.88$                                     & $0.78$                                     \\
                                                                & GPT-$3.5$                                & $0.90$                                     & $0.88$                                     & $0.91$                                     & $0.79$                                     & $0.89$                                     & $0.87$                                     & $0.90$                                     & $0.79$                                     \\
                                                                & Presidio                               & $0.91$                                     & $0.86$                                     & $0.89$                                     & $0.82$                                     & $0.87$                                     & $0.85$                                     & $0.91$                                     & $0.78$                                         \\
                                                                & FLAIR                                  & $0.90$                                     & $0.87$                                     & $0.88$                                     & $0.80$                                     & $0.90$                                      & $0.88$                                     & $0.88$                                     & $0.80$                                     \\
                                                                & NER CoNLL$2003$ BERT                     & $0.89$                                     & $0.87$                                     & $0.89$                                     & $0.77$                                     & $0.91$                                     & $0.85$                                     & $0.90$                                     & $0.81$                                     \\ \hline
\end{tabular}}
\caption{Performance of Task $1$ (Occupancy Prediction) for the six NER models. The results are categorized depending on the private information removed: locations (LOC) and persons (PER). This experiment incorporates the use of two pre-trained Transformers, which optimize content understanding and enhance the models' effectiveness.}
\label{tab:scoringtool}
\end{table*}
\subsection{Privacy-aware Models: Impact of Anonymized Text}\label{subsec:privacytool}

The experimental framework illustrates the functionality of a Recruitment Tool using anonymized data, ensuring the protection of sensitive information. This approach enables an effective evaluation of the tool’s ability to classify candidates and assess gender bias in recruitment processes. Anonymized resumes are used to preserve privacy while retaining the realism of data.

The resume scoring tool is initially trained using textual data from non-anonymized resumes, establishing a baseline performance benchmark. This baseline serves as a reference point to assess how the model performs when all personal details are available, allowing for a direct comparison in later trials. Following this, the resume scoring tool is trained on anonymized data using six independent NER modules. Each NER module masked the same entity types (names and locations), and the goal is to determine whether the removal of these entities affects the performance of the scoring tool relative to the established baseline (i.e., non-anonymized data).

\begin{figure}[tp]
\centering
\includegraphics[width=\textwidth]{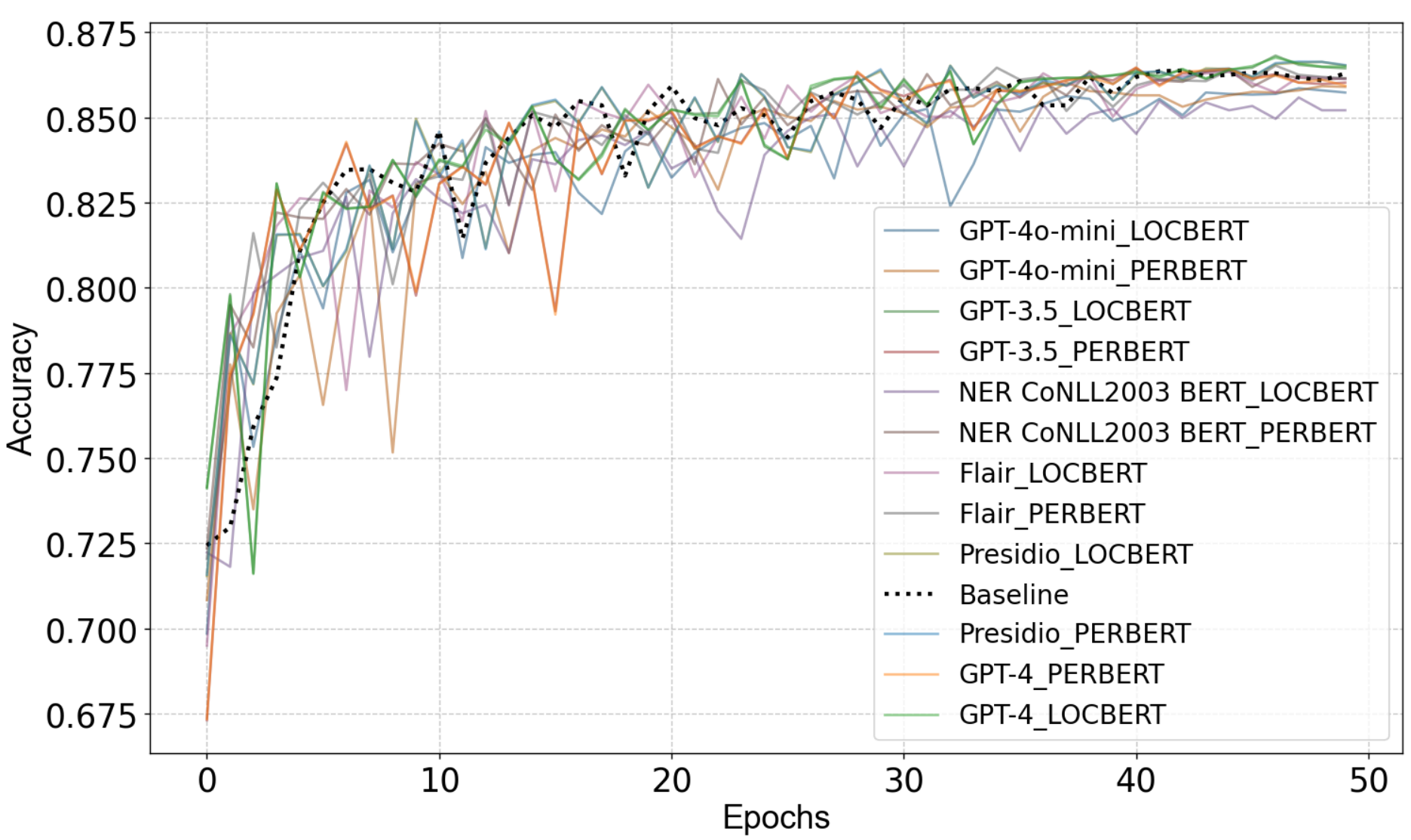}
\caption{Accuracy over epochs for the BERT-based occupancy prediction model. The Figure compares different implementations, including six anonymization modules over different two entities (PER and LOC). The baseline model, without anonymization module, is highlighted with a dotted black line. The results reflect performance across different configurations, with darker colors indicating specific models.}
\label{fig:accbert}
\end{figure}

Our experiments include two learning tasks (occupancy prediction and scoring prediction). The results obtained for the occupancy prediction are summarized in Table \ref{tab:scoringtool}, which outlines the comparative performance metrics for each anonymization model, highlighting the extent to which the anonymization process impacts the performance and effectiveness of the resume scoring tool.
As reflected in Table \ref{tab:scoringtool}, the results obtained compared to the baseline are even slightly superior, especially in cases where the anonymization method has been conducted with a GPT-based model. This shows that these methods are capable of, at the very least, maintaining performance and, in some cases, enhancing privacy. These findings suggest that it is possible to strike a balance between protecting privacy and achieving good model performance. In the worst case, performance remains unchanged, while in the best case, privacy is increased without compromising accuracy, which is a significant benefit when dealing with sensitive data.
\begin{figure}[t!]
\centering
\includegraphics[width=\textwidth]{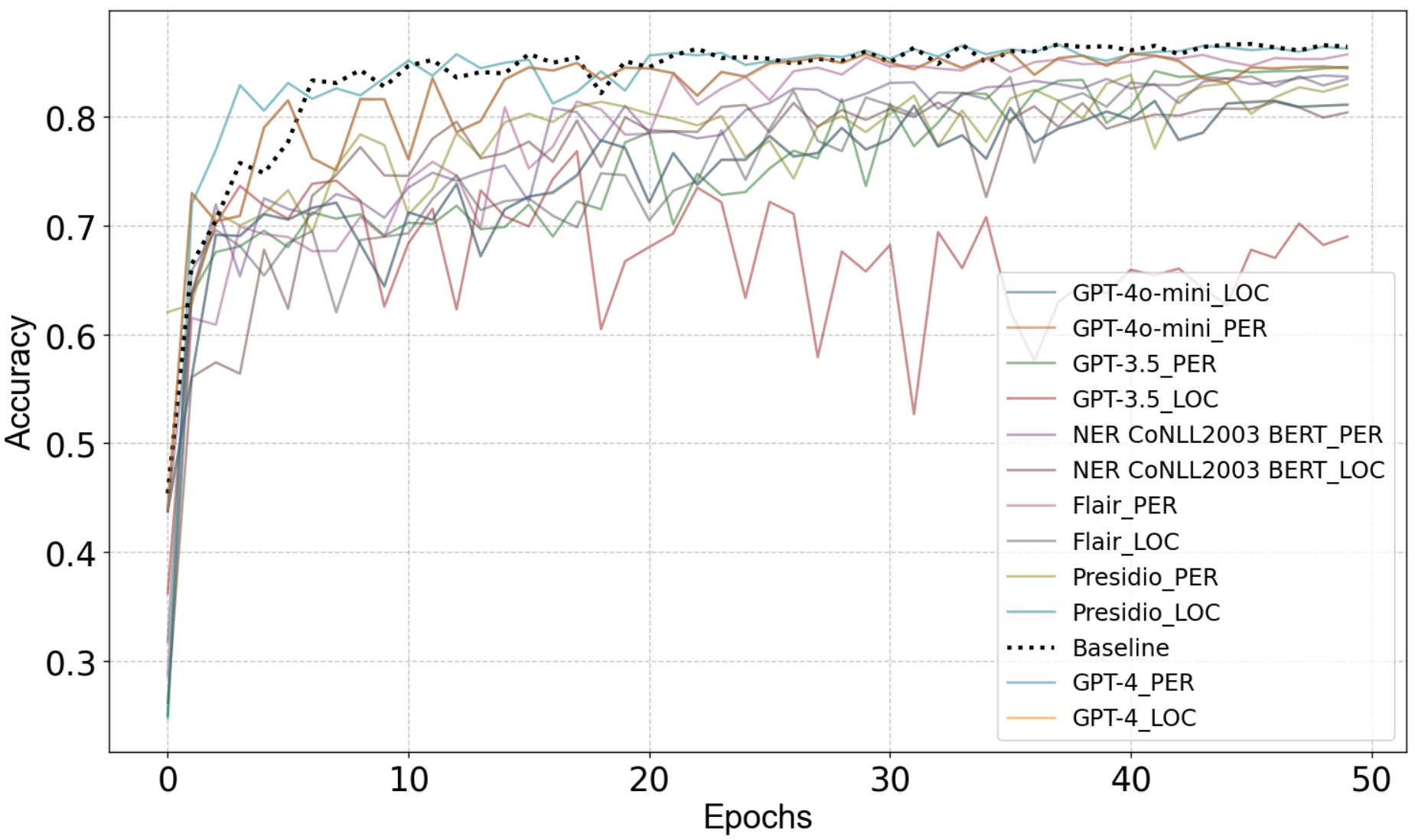}
\caption{Accuracy over epochs for the RoBERTa-based occupancy prediction model. The figure compares different implementations, including six anonymization modules over two different entities (PER and LOC). The baseline model, without anonymization module, is highlighted with a dotted black line. The results reflect performance across different configurations, with darker colors indicating specific models.}
\label{fig:accroberta}
\end{figure}
Furthermore, the learning process of the models throughout these experiments is depicted in Fig.~\ref{fig:accbert} for BERT and Fig.~\ref{fig:accroberta} for RoBERTa, where it becomes clear that the learning trajectory of the model is influenced both by the anonymization technique and the specific type of entity being masked. This figure provides valuable information on how the model adapts to varying levels of information removal, showing whether certain anonymization methods lead to better generalization or, conversely, hinder the model’s ability to maintain consistent scoring performance. Ultimately, this experiment sheds light on the delicate balance between the preservation of individual privacy and the predictive power of machine learning models in sensitive domains.
\begin{table*}[tp]
\centering
\resizebox{\textwidth}{!}{
\begin{tabular}{llcccccccc}
\hline
\multicolumn{2}{l}{} & \multicolumn{4}{c}{\textbf{LOC}} & \multicolumn{4}{c}{\textbf{PER}} \\
\textbf{Transformer} & \textbf{Anonymization Module} & \textbf{O1} & \textbf{O2} & \textbf{O3} & \textbf{O4} & \textbf{O1} & \textbf{O2} & \textbf{O3} & \textbf{O4} \\
\hline
\multirow{8}{*}{RoBERTa}
& \textbf{None (Baseline)}          & 0.86 & 0.88 & 0.85 & 0.79 & 0.89 & 0.84 & 0.87 & 0.72 \\
&\textbf{None (Bias-aware)}& 0.88 & 0.84 & 0.85 & 0.82 & 0.91 & 0.87 & 0.86 & 0.75 \\
& GPT-4                             & 0.91 & 0.95 & 0.84 & 0.82 & 0.89 & 0.87 & 0.85 & 0.75 \\
& GPT-4o-mini                       & 0.89 & 0.90 & 0.86 & 0.80 & 0.92 & 0.85 & 0.87 & 0.64 \\
& GPT-3.5                           & 0.82 & 0.85 & 0.85 & 0.65 & 0.90 & 0.87 & 0.90 & 0.73 \\
& Presidio                          & 0.90 & 0.90 & 0.89 & 0.78 & 0.91 & 0.83 & 0.84 & 0.71 \\
& FLAIR                             & 0.90 & 0.86 & 0.83 & 0.74 & 0.89 & 0.85 & 0.87 & 0.76 \\
& NER CoNLL2003 BERT                & 0.94 & 0.85 & 0.83 & 0.69 & 0.86 & 0.89 & 0.91 & 0.80 \\
\hline
\multirow{8}{*}{BERT}
& \textbf{None (Baseline)}          & 0.87 & 0.85 & 0.87 & 0.79 & 0.89 & 0.86 & 0.87 & 0.79 \\
&\textbf {None (Bias-aware)} & 0.89 & 0.86 & 0.90 & 0.83 & 0.89 & 0.86 & 0.92 & 0.77 \\
& GPT-4                             & 0.94 & 0.87 & 0.87 & 0.82 & 0.94 & 0.92 & 0.91 & 0.79 \\
& GPT-4o-mini                       & 0.93 & 0.87 & 0.91 & 0.77 & 0.90 & 0.91 & 0.89 & 0.83 \\
& GPT-3.5                           & 0.88 & 0.90 & 0.89 & 0.75 & 0.88 & 0.85 & 0.92 & 0.81 \\
& Presidio                          & 0.89 & 0.88 & 0.93 & 0.81 & 0.85 & 0.89 & 0.90 & 0.74 \\
& FLAIR                             & 0.92 & 0.86 & 0.88 & 0.81 & 0.91 & 0.88 & 0.90 & 0.77 \\
& NER CoNLL2003 BERT                & 0.91 & 0.88 & 0.90 & 0.75 & 0.93 & 0.86 & 0.91 & 0.82 \\
\hline
\end{tabular}
}
\caption{Performance of Occupancy Prediction for the six NER models. The results are categorized depending on the private information removed: locations (LOC), persons (PER) and Gender (all rows except \textbf{None (Baseline)}). This experiment incorporates the use of two pre-trained Transformers, which optimize content understanding and enhance the models' effectiveness.}
\label{tab:biasaware}
\end{table*}

\subsection{Privacy- and Bias-aware Models}
In Section \ref{subsec:privacytool} it has been observed that anonymizing text related to PER (persons) and LOC (locations) does not impact the performance of the occupation prediction model. Building on this insight, the objective is to create a recruitment tool that prioritizes both privacy and fairness, ensuring that it is free of gender bias.

The approach involves several steps. First, entity anonymization will be implemented to mask sensitive personal data, such as names and locations. This ensures that no identification information influences the decision-making process. Second, the elimination of gender bias will target both explicit indicators (e.g., pronouns, titles) as in Section \ref{subsec:privacytool} and implicit markers (e.g., gendered phrases or role associations). To address this, we will employ the bias-aware learning framework introduced in \cite{pena2023human} and further detailed in \cite{peña2025bias}, which represents a method to systematically remove gender characteristics from the model's internal representations. This approach allows the model to make predictions that are independent of gender-specific information while preserving task-relevant linguistic patterns, ensuring that gender does not influence predictions. In combination, these steps aim to produce a recruitment tool that is both privacy conscious and bias aware.

As can be seen in Table \ref{tab:biasaware}, the results demonstrate that the combined application of anonymization of entities (masking private information such as locations and persons) along with gender bias mitigation techniques does not degrade the performance of occupation prediction models. Rather, this dual approach often results in stable or improved predictive accuracy compared to baseline models without any anonymization or bias-aware learning.  Importantly, the findings confirm that the implementation of privacy-preserving and fairness-enhancing mechanisms together can enhance the robustness and generalization capacity of the models without sacrificing performance.

The results shown before have been evaluated for the task of Occupancy Classification (see Sect.\ref{sec:tasks}). We now evaluate our PBa-LLM approach (incorporating both NER-based privacy enhancement and gender bias reduction, similar to Table\ref{tab:biasaware}) for the other task studied in the present paper: Candidate Scoring. For this second task of candidate scoring, we follow the same protocol of \cite{pena2023human,peña2025bias}, generating a top 100 list of candidates based on the estimated candidate scores $\hat{y}_i$ (see the right output in Fig.\ref{fig:architecture}). Recall that we used the men-biased dataset from FairCVdb, so the top 100 list is expected to have more men than women for similar competencies given the gender-biased expert score annotations. Eliminating personal data and gender cues in biographies using our PBa-LLMs, Table~\ref{tab:combined_bias} shows that even with these gender-biased training data, one can generate privacy-preserving bias-free AI models (useful enough for NLP tasks like the two tasks studied here: Occupancy Classification and Candidate Scoring).

\begin{table*}[h!]
\centering
\begingroup
\large
\resizebox{\textwidth}{!}{%
\begin{tabular}{llcccc}
\hline
\textbf{Transformer} & \textbf{Anonymization Module} 
& \multicolumn{2}{c}{\textbf{LOC}} 
& \multicolumn{2}{c}{\textbf{PER}} \\

& & \textbf{Prop (M)} & \textbf{Prop (F)} 
  & \textbf{Prop (M)} & \textbf{Prop (F)} \\ \hline

\multirow{8}{*}{RoBERTa}                                    
& \textbf{None (Baseline)}            & $67.00\%$ & $33.00\%$ & $67.00\%$ & $33.00\%$ \\
& \textbf{None (Bias-aware)} & $47.80\%$ & $52.20\%$ & $47.80\%$ & $52.20\%$ \\
& GPT-$4$                                      & $50.60\%$ & $49.40\%$ & $50.60\%$ & $49.40\%$ \\
& GPT-$4$o-mini                                & $50.80\%$ & $49.20\%$ & $50.90\%$ & $49.10\%$ \\
& GPT-$3.5$                                    & $50.90\%$ & $49.10\%$ & $50.70\%$ & $49.30\%$ \\
& Presidio                                     & $50.70\%$ & $49.30\%$ & $50.80\%$ & $49.20\%$ \\
& FLAIR                                        & $50.75\%$ & $49.25\%$ & $50.85\%$ & $49.15\%$ \\
& NER CoNLL$2003$ BERT                         & $50.80\%$ & $49.20\%$ & $50.75\%$ & $49.25\%$ \\ \hline

\multirow{8}{*}{BERT}                                       
& \textbf{None (Baseline)}            & $69.40\%$ & $30.60\%$ & $69.40\%$ & $30.60\%$ \\
& \textbf{None (Bias-aware)} & $48.60\%$ & $51.40\%$ & $48.60\%$ & $51.40\%$ \\
& GPT-$4$                                      & $50.70\%$ & $49.30\%$ & $50.85\%$ & $49.15\%$ \\
& GPT-$4$o-mini                                & $50.80\%$ & $49.20\%$ & $50.95\%$ & $49.05\%$ \\
& GPT-$3.5$                                    & $50.70\%$ & $49.30\%$ & $50.80\%$ & $49.20\%$ \\
& Presidio                                     & $50.75\%$ & $49.25\%$ & $50.88\%$ & $49.12\%$ \\
& FLAIR                                        & $50.75\%$ & $49.25\%$ & $50.92\%$ & $49.08\%$ \\
& NER CoNLL$2003$ BERT                         & $50.80\%$ & $49.20\%$ & $50.83\%$ & $49.17\%$ \\ \hline
\end{tabular}%
}
\endgroup
\caption{
Gender proportions in the FairCVdb validation set for both LOC and PER labels were analyzed using candidate shortlists from resume scores $\hat{y}$ across different scoring tool configurations. This table reports the male and female representation to assess gender balance after different anonymization strategies.
}
\label{tab:combined_bias}
\end{table*}

\section{Conclusions}
\label{sec:conclusions}

In this work, a privacy-preserving learning framework was proposed and evaluated using Named Entity Recognition (NER) to anonymize sensitive information in text while maintaining system performance. The framework demonstrated its effectiveness in automated recruitment tools using two Transformer-based language models and six anonymization algorithms on a dataset of 24,000 resumes. The experimental results revealed that anonymized data preserved the utility of the models for tasks such as occupancy and scoring predictions, while simultaneously enhancing privacy and mitigating bias. These findings underscore the potential of NER-based anonymization techniques as a scalable and domain-independent solution to ensure both ethical compliance and trust in AI-driven decision-making systems. 

Furthermore, the study highlighted the robustness of the NER-based approach, which showed minimal degradation between different anonymization models. The results also revealed that, in certain scenarios, privacy-preserving techniques not only safeguarded sensitive information but also led to slight improvements in fairness, as seen in bias-aware recruitment tasks.

\section{Future Works}
A promising direction for future work involves extending the current NER-based anonymization framework to other sensitive domains beyond recruitment, such as healthcare \cite{ai4food}, education \cite{becerra2025aibased}, or legal affairs \cite{PENA-Layout}, where privacy concerns are critical. These areas introduce different entity types and context-dependent privacy risks, necessitating the development of domain-specific NER models or hybrid anonymization strategies that combine rule-based and learned approaches. In addition, embedding adaptive mechanisms that automatically calibrate the strength of anonymization to the sensitivity of each domain would help balance privacy with data utility in real-time deployments.

Another important avenue is to expand the bias mitigation capabilities of the system beyond gender to address other social and demographic biases \cite{2021_TTS_Biases_Terhorst}, such as ethnicity, age, disability, or socioeconomic status. Incorporating detection and debiasing techniques \cite{2022_SafeAI_IFBiD_Serna,2023_ECAIw_LFIT-XAI_Tello,2023_COMPSAC_BiasAI_N-sigma_DeAlcala,peña2025bias} tailored to these attributes would promote fairness on multiple axes and make the recruitment tool more inclusive and equitable across diverse candidate populations.

\section{Acknowledgement}
This study has been supported by projects HumanCAIC (TED2021-131787B-I00 MICINN), Cátedra ENIA UAM-VERIDAS en IA Responsable (NextGenerationEU PRTR TSI-100927-2023-2), and R\&D Agreement DGGC/UAM/FUAM for Biometrics and Applied AI. G. Mancera is supported by FPI-PRE2022-104499 MICINN/FEDER. Work conducted in the ELLIS Unit Madrid.
%
%
%
\bibliographystyle{splncs04}
\bibliography{mybibliography}

\end{document}